%% file: main.tex
\documentclass[sigconf]{acmart}
\AtBeginDocument{%
  }

\setcopyright{none}           
\renewcommand\footnotetextcopyrightpermission[1]{} 
\author{Arun Vignesh Malarkkan}
\email{arun.malarkkan@asu.edu}
\affiliation{%
  \institution{Arizona State University}
  \city{Tempe}
  \state{Arizona}
  \country{USA}
}

\author{Xinyuan Wang}
\email{xwang735@asu.edu}
\affiliation{%
  \institution{Arizona State University}
  \city{Tempe}
  \state{Arizona}
  \country{USA}
}

\author{Kunpeng Liu}
\email{kunpenl@clemson.edu}
\affiliation{%
  \institution{Clemson University}
  \city{Clemson}
  \state{South Carolina}
  \country{USA}
}

\author{Denghui Zhang}
\email{dzhang42@stevens.edu}
\affiliation{%
  \institution{Stevens Institute of Technology}
  \city{Hoboken}
  \state{New Jersey}
  \country{USA}
}

\author{Yanjie Fu}
\email{yanjie.fu@asu.edu}
\affiliation{%
  \institution{Arizona State University}
  \city{Tempe}
  \state{Arizona}
  \country{USA}
}

\usepackage{times}
\usepackage{latexsym}
\usepackage{amsmath}
\usepackage{colortbl}
\usepackage{pifont}
\usepackage{diagbox}
\usepackage{tablefootnote}
\usepackage{threeparttable}

\usepackage{hyperref}
\usepackage{url}
\usepackage{graphicx}
\usepackage{verbatim}
\usepackage{fancyvrb}
\usepackage{listings}
\usepackage{booktabs}
\usepackage{array}  
\usepackage{xcolor}   
\usepackage{cleveref}

\usepackage{array}
\usepackage{booktabs}
\usepackage{longtable}
\usepackage{multirow}
\usepackage{tabularx}
\usepackage{ragged2e}
\usepackage{makecell}
\usepackage{multirow}
\usepackage{amsmath}
\usepackage{array}
\usepackage{booktabs}
\usepackage{enumitem}
\usepackage{bbm}
\usepackage{subfigure}

\begin{document}

\title{Causally-Guided Diffusion for Stable Feature Selection}

\input{sections/abstract}


\maketitle

\input{sections/introduction}
\input{sections/problem_statement}
\input{sections/method}

\input{sections/experiments}
\input{sections/related_works}
\input{sections/conclusion}

%

\bibliographystyle{ACM-Reference-Format}
\bibliography{main}

\newpage
\appendix

\input{sections/appendix}

\end{document}

%% file: sections/abstract.tex

\begin{abstract}
Feature selection is fundamental to robust data-centric AI, but most existing methods optimize predictive performance under a single data distribution. This often selects spurious features that fail under distribution shifts. Motivated by principles from causal invariance, we study feature selection from a stability perspective and introduce \emph{Causally-Guided Diffusion for Stable Feature Selection} (CGDFS). In CGDFS, we formalized feature selection as approximate posterior inference over feature subsets, whose posterior mass favors low prediction error and low cross-environment variance. Our framework combines three key insights: First, we formulate feature selection as stability-aware posterior sampling. Here, causal invariance serves as a soft inductive bias rather than explicit causal discovery. Second, we train a diffusion model as a learned prior over plausible continuous selection masks, combined with a stability-aware likelihood that rewards invariance across environments. This diffusion prior captures structural dependencies among features and enables scalable exploration of the combinatorially large selection space. Third, we perform guided annealed Langevin sampling that combines the diffusion prior with the stability objective, which yields a tractable, uncertainty-aware posterior inference that avoids discrete optimization and produces robust feature selections.
We evaluate CGDFS on open-source real-world datasets exhibiting distribution shifts. Across both classification and regression tasks, CGDFS consistently selects more stable and transferable feature subsets, which leads to improved out-of-distribution performance and greater selection robustness compared to sparsity-based, tree-based, and stability-selection baselines.

\end{abstract}

%% file: sections/introduction.tex

\section{Introduction}

Feature selection is fundamental to building robust data-centric machine learning systems, aiming to identify compact subsets of features that support efficient, interpretable, and generalizable models. However, as models are increasingly deployed in dynamic, real-world environments, a critical challenge emerges: \textbf{existing feature selection methods systematically select spurious features that fail under distribution shift}. Consider a credit risk model trained on historical data, where ``home ownership'' spuriously correlates with default rates due to regional economic conditions. Standard feature selection methods will select this feature based on its strong in-distribution correlation, yet the model fails when deployed to new markets where this correlation is not valid. This brittleness poses severe risks in high-stakes applications spanning healthcare, finance, and algorithmic fairness.

Despite extensive study, most feature selection methods remain focused on correlation. Filter methods such as mutual information and correlation-based criteria~\cite{yu2003feature, hoque2014mifs}, wrapper methods including forward/backward selection~\cite{el2016review}, and embedded methods such as LASSO and elastic net~\cite{muthukrishnan2016lasso, zou2005regularization} all optimize predictive performance under a single training distribution. While effective at identifying features correlated with the target, these methods fail to distinguish features whose predictive utility remains stable across environments from spurious features that are predictive only due to distribution-specific correlations. This observation exposes a fundamental limitation: \textbf{feature selection methods that rely solely on in-distribution performance provide no robustness guarantees under distribution shift}.

Insights from Causal learning offer a principled direction for addressing this limitation through the notion of \emph{invariance}: predictive relationships that remain stable across environments are more likely to generalize, whereas distribution-specific correlations tend to break under shift~\cite{buhlmann2020invariance, malarkkan2025rethinking}. Several methods, including Invariant Risk Minimization (IRM)~\cite{arjovsky2019invariant} and Invariant Causal Prediction (ICP)~\cite{peters2016causal}, operationalize this principle by enforcing stability across multiple environments. However, directly translating invariance-based ideas to feature selection presents substantial challenges.
1) Stability-driven objectives typically induce highly non-convex optimization landscapes, in which gradient-based or greedy methods are prone to poor local optima. 
2) Feature selection is inherently combinatorial, requiring the selection of a discrete subset from an exponentially large space of $2^p$ possible feature sets, which makes standard continuous optimization techniques difficult to apply. 
Together, these challenges have limited the practical use of causal invariance-based principles for robust feature selection.

\noindent\textbf{Our research perspectives: }Based on these observations, we reframe feature selection under distribution shift. First, stability across environments should be treated as a primary criterion for feature relevance, rather than a secondary diagnostic applied after selection. Features that yield consistently strong predictive behavior across heterogeneous environments are more likely to generalize beyond the training distribution. Second, feature relevance under distribution shift is inherently uncertain: multiple distinct feature subsets may exhibit comparable stability and predictive performance, particularly in the presence of correlated or redundant features. Third, this uncertainty in feature selection is combinatorial, rendering deterministic optimization strategies ill-suited for capturing the full space of plausible feature sets. These insights motivate feature selection formulations that reason over distributions of feature subsets rather than committing to a single optimum.

Based on this perspective, we propose \textbf{Causally-Guided Diffusion for Stable Feature Selection (CGDFS)}, a principled framework that formulates feature selection as approximate posterior inference over feature subsets. Rather than searching for a single optimal subset via greedy search or continuous relaxation, CGDFS samples from a posterior distribution that favors feature subsets that achieve both strong predictive performance and stable behavior across multiple environments. This probabilistic formulation operationalizes the causal invariance-based principles as soft inductive bias without requiring interventional data or explicit causal graph discovery. By reasoning over distributions of feature subsets, CGDFS naturally captures uncertainty, accommodates multiple plausible solutions, and avoids the brittleness of deterministic feature selection strategies.

CGDFS operationalizes this probabilistic framework with three components. First, we represent feature subsets as continuous selection masks in $[0,1]^p$, and a diffusion model is trained to learn a prior over plausible subsets. This learned prior captures global structure in the feature space, including feature complementarity and redundancy, and enables efficient exploration of the combinatorially large subset space. Second, we define a stability-aware objective that evaluates feature subsets based on both predictive performance and cross-environment variability, favoring subsets whose predictive behavior remains consistent across environments. Third, we perform guided diffusion sampling using annealed Langevin dynamics, combining gradients from the learned diffusion prior with gradients from the stability objective to approximate posterior inference over feature subsets. This sampling-based formulation avoids discrete combinatorial optimization while naturally capturing posterior uncertainty.
In summary, this paper makes the following contributions:
\begin{itemize}
    \item We formulate feature selection under distribution shift as approximate posterior inference over feature subsets, guided by a stability-based objective inspired by causal invariance principles.
    \item We introduce a diffusion-based prior over continuous feature selection masks, and a guided sampling procedure that enables scalable, uncertainty-aware exploration of the combinatorial feature subset space without discrete search or greedy optimization.
    \item Through extensive experiments on synthetic benchmarks and real-world tabular datasets with distribution shifts, we demonstrate that CGDFS consistently selects more stable and transferable feature subsets, leading to improved out-of-distribution predictive performance compared to competitive baselines.
\end{itemize}

%% file: sections/problem_statement.tex

\section{Problem Statement}

We consider supervised learning on tabular data under distribution shift. 
Let $\mathcal{D} = \bigcup_{e=1}^E \mathcal{D}^{(e)}$ denote a dataset partitioned into
$E$ environments, where
$\mathcal{D}^{(e)} = \{(x_i^{(e)}, y_i^{(e)})\}_{i=1}^{n_e}$,
$x_i^{(e)} \in \mathbb{R}^p$, and $y_i^{(e)} \in \mathcal{Y}$.
Let $z \in \{0,1\}^p$ denote a feature selection vector, and let
$f_\theta(\cdot)$ be a predictive model parameterized by $\theta$.

For a given feature subset $z$, we define the environment-specific expected loss
\[
\mathcal{L}^{(e)}(z)
= \mathbb{E}_{(x,y)\sim \mathcal{D}^{(e)}} \big[ \ell\big(f_\theta(x \odot z), y\big) \big],
\]
where $\ell(\cdot,\cdot)$ denotes a task-appropriate loss function.
We summarize predictive performance across environments by the mean loss
\[
\bar{\mathcal{L}}(z)
= \frac{1}{E} \sum_{e=1}^E \mathcal{L}^{(e)}(z),
\]
and quantify robustness under distribution shift by the loss variance
\[
\mathrm{Var}_e\big(\mathcal{L}^{(e)}(z)\big)
= \frac{1}{E} \sum_{e=1}^E
\big(\mathcal{L}^{(e)}(z) - \bar{\mathcal{L}}(z)\big)^2.
\]

The feature selection problem under distribution shift is to identify feature subsets
$z$ that jointly achieve low average predictive loss and low cross-environment
performance variability. Formally, this defines a multi-objective optimization problem
over the combinatorial space $\{0,1\}^p$:
\[
\min_{z \in \{0,1\}^p}
\quad
\Big( \bar{\mathcal{L}}(z), \; \mathrm{Var}(\mathcal{L}(z)) \Big).
\]

%% file: sections/method.tex

\section{Methodology}

We present \textbf{Causally-Guided Diffusion for Stable Feature Selection (CGDFS)}, a framework that formulates feature selection under distribution shift as approximate posterior inference over feature subsets.CGDFS combines a diffusion-based prior over feature selection masks with a stability-aware objective, enabling scalable inference in high-dimensional and combinatorial feature spaces without discrete search.
In this section, we first introduce the feature subset representation and modeling assumptions (\S\ref{subsec:problem}). We then describe the three core components of CGDFS: diffusion-based prior learning (\S\ref{subsec:prior}), stability-based likelihood construction (\S\ref{subsec:likelihood}), and guided posterior sampling via annealed Langevin dynamics (\S\ref{subsec:sampling}).

\begin{figure*}[t]
  \centering
  \includegraphics[height=6.3cm, width=0.96\textwidth]{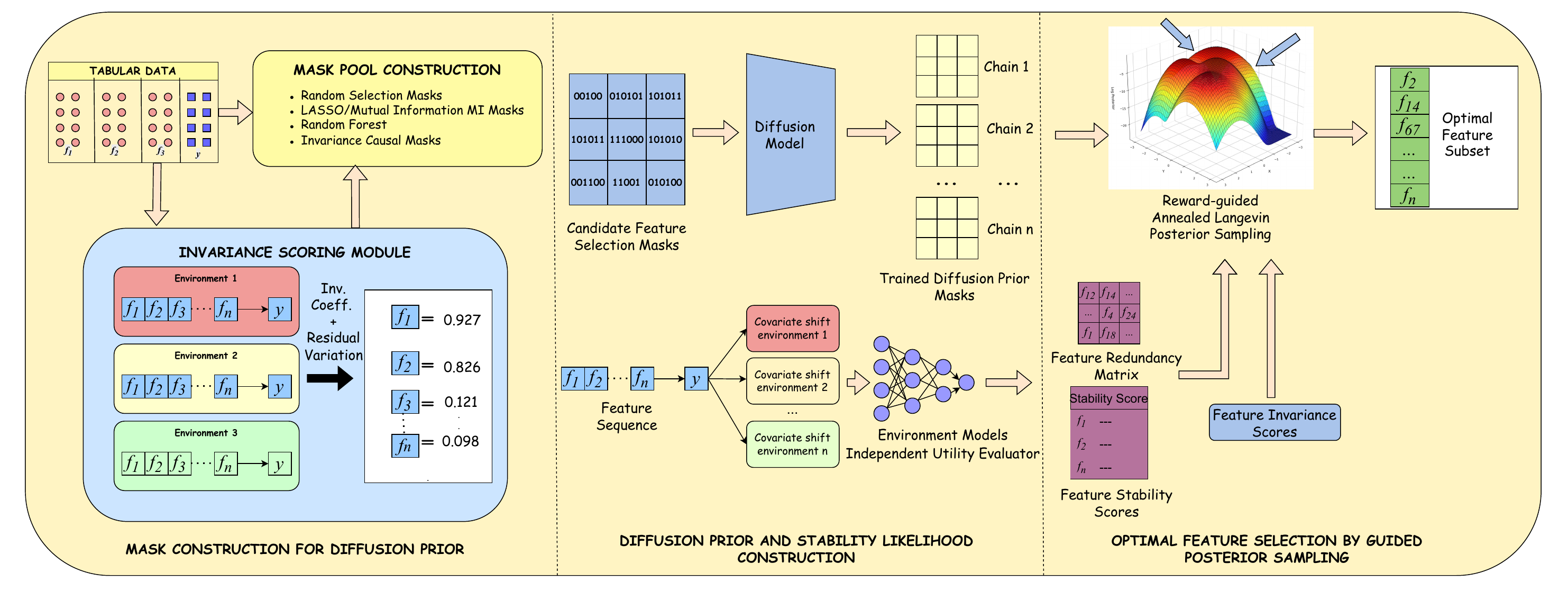}
  \caption{Framework overview of CGDFS. Feature selection is formulated as approximate posterior inference over feature subsets. A diffusion model learns a structured prior over continuous feature masks from a diverse mask pool. A stability-based likelihood evaluates predictive performance and cross-environment variability. Guided Langevin dynamics combines prior and likelihood gradients to sample from the posterior, and the samples are aggregated to produce stable, interpretable feature selections, with invariance-based causal signals incorporated as soft guidance through the prior.}
  \vspace{-0.3cm}
  \label{fig:framework}
\end{figure*}

\subsection{Feature Subset Representation and Modeling Assumptions}
\label{subsec:problem}

We represent a feature subset by a binary selection vector $z \in \{0,1\}^p$, where $z_j = 1$ indicates that feature $j$ is selected. Direct optimization over this discrete space is computationally intractable due to its exponential size. To enable scalable inference, we relax the discrete selection vector to a continuous representation and model feature subsets using selection masks
\[
z \in [0,1]^p,
\]
where each coordinate $z_j$ represents a soft selection weight for feature $j$. Given a predictive model $f_\theta$, the mask induces a masked input $x \odot z$, where $\odot$ denotes element-wise multiplication. This relaxation preserves the semantics of feature selection while enabling optimization and inference in continuous space.

\noindent\underline{\emph{Environment-specific evaluation.}}
We assume access to a finite set of environments $\{\mathcal{D}^{(e)}\}_{e=1}^E$ during training, which are explicitly constructed or observed to exhibit covariate shifts. Each environment corresponds to a distinct data distribution under which predictive performance may vary. For a given selection mask $z$, performance is evaluated separately within each environment using masked inputs. While environments are designed to induce distributional heterogeneity, we do not assume access to interventional data or knowledge of the underlying causal graph; environments are used as observational settings for assessing stability under covariate shift.

\noindent\underline{\emph{Modeling assumptions.}}
Our formulation relies on two small assumptions. First, the predictive model $f_\theta$ is differentiable with respect to its input, allowing gradients to propagate through the feature mask $z$. Second, predictive performance within each environment can be summarized by an environment-specific loss function.

\subsection{Learning a Diffusion-Based Prior}
\label{subsec:prior}

We begin by learning a prior distribution over feature subsets that captures structural regularities among plausible selections. Feature subsets often exhibit non-trivial dependencies, including complementarity: features that are informative only when selected jointly and redundancy: groups of features that provide overlapping information. Ignoring such structures leads to inefficient exploration of the combinatorial subset space and unstable selection behavior. Rather than imposing a hand-crafted prior, we learn a data-driven prior from empirically generated feature subset candidates.

\noindent\underline{\emph{Mask pool construction}}
To approximate the unknown distribution over plausible feature subsets, we construct a diverse pool of candidate masks
$\mathcal{M} = \{ s^{(1)}, \ldots, s^{(M)} \} \subset [0,1]^p$.
The mask pool is generated with multiple complementary heuristics, including random feature selection, sparsity-inducing linear models, information-theoretic ranking, and tree-based feature importance methods. This combination introduces distinct inductive biases and collectively populates both sparse and dense regions of the subset space.
Each candidate subset is represented as a continuous selection vector in $[0,1]^p$, obtained by adding small Gaussian perturbations to binary selections and clipping to the unit interval. This smoothing step avoids degenerate binary support and facilitates diffusion model training while preserving the underlying subset structure.

\noindent\underline{\emph{Diffusion-based prior.}}
We model the distribution over feature masks using a score-based diffusion framework. Let $q(s^{(0)})$ denote the empirical measure induced by the mask pool. A forward diffusion process progressively perturbs an initial mask $s^{(0)}$ by injecting Gaussian noise over a sequence of timesteps, yielding noisy representations $\{s^{(t)}\}_{t=1}^T$:
\begin{equation}
q(s^{(t)} \mid s^{(0)}) =
\mathcal{N}\!\left(
s^{(t)}; \sqrt{\bar{\alpha}_t} \, s^{(0)}, (1 - \bar{\alpha}_t) I
\right),
\end{equation}
where $\{\bar{\alpha}_t\}$ is defined by a standard noise schedule.

To learn the diffusion-based prior, a parameterized score function $s_\theta(s^{(t)}, t)$ is trained to approximate the reverse-time dynamics by predicting the injected noise at each timestep. Training is performed via the standard denoising score-matching objective:
\begin{equation}
\mathcal{L}_{\mathrm{prior}}(\theta)
=
\mathbb{E}_{s^{(0)}, t, \epsilon}
\left[
\left\| s_\theta(s^{(t)}, t) - \epsilon \right\|^2
\right],
\label{eq:diffusion_loss}
\end{equation}
where
$s^{(t)} = \sqrt{\bar{\alpha}_t} s^{(0)} + \sqrt{1 - \bar{\alpha}_t} \, \epsilon$,
$t$ is sampled uniformly from $\{1,\ldots,T\}$, and
$\epsilon \sim \mathcal{N}(0, I)$.

After training, the diffusion model implicitly defines a prior distribution $p(s)$ over feature selection masks. The learned score function $s_\theta(s,t)$ provides gradients of the log-density of this prior, characterizing the geometry of high-probability regions. This prior assigns higher probability to feature configurations reflecting structured dependencies observed in the mask pool, without explicit enumeration of subsets or manual specification of feature interactions. This learned prior later guides posterior inference toward plausible and structured feature subsets.

\vspace{-0.2cm}
\subsection{Stability-Based Likelihood}
\label{subsec:likelihood}

We next define a likelihood function that evaluates feature subsets based on both predictive performance and robustness across environments. This likelihood provides the mechanism through which causal invariance-inspired behavior is enforced in CGDFS.

\noindent\underline{\emph{Environment-specific evaluation.}}
Given a continuous feature selection mask $s \in [0,1]^p$, predictive performance is evaluated separately within each environment.
For each environment $e \in \{1,\ldots,E\}$, we train a lightweight predictor
$f_{\phi_e} : \mathbb{R}^p \rightarrow \mathcal{Y}$
using masked inputs $x \odot s$.
Training separate predictors avoids parameter sharing across environments and ensures that stability is assessed through consistency of predictive behavior rather than through shared model weights.
To mitigate overfitting on limited per-environment data, the predictor architecture is intentionally simple.
Training is performed via empirical risk minimization on a training split of environment $e$, after which the predictor parameters are frozen.
Once trained, the frozen predictors $\{f_{\phi_e^*}\}_{e=1}^E$ are used to evaluate feature subsets on held-out validation data.
Crucially, although predictor parameters remain fixed, the loss remains differentiable with respect to the feature mask $s$,
allowing gradients to propagate through the masked input during posterior inference.

\noindent\underline{\emph{Performance and stability criteria}}
For a given mask $s$, let
\[
\ell_e(s)
=
\mathbb{E}_{(x,y)\sim\mathcal{D}^{(e)}_{\mathrm{val}}}
\big[\ell(f_{\phi_e^*}(x \odot s), y)\big]
\]
denote the validation loss on environment $e$.
We summarize predictive performance across environments using the mean loss
\[
\bar{\ell}(s)
=
\frac{1}{E}\sum_{e=1}^E \ell_e(s),
\]
and quantify robustness under distribution shift using the variance
\[
\mathrm{Var}_e\!\big(\ell_e(s)\big)
=
\frac{1}{E}\sum_{e=1}^E
\big(\ell_e(s) - \bar{\ell}(s)\big)^2.
\]
Feature subsets that achieve low mean loss but high variance are predictive yet unstable, whereas subsets with low values of both terms exhibit strong and consistent performance across environments.

\noindent\underline{\emph{Stability-based Likelihood.}}
We combine both performance and cross-environment variability into a single stability-based objective function
\[
U(s)
=
\bar{\ell}(s)
+
\lambda_{\mathrm{var}}\,\mathrm{Var}_e\!\big(\ell_e(s)\big),
\]
where $\lambda_{\mathrm{var}}\ge 0$ controls the trade-off between between accuracy and stability.
Lower objective values correspond to feature subsets that achieve favorable performance-robustness trade-offs.
We define the likelihood over feature subsets implicitly via this objective as
\[
\log p(\mathcal{D}\mid s)\;\propto\; -U(s).
\]
does not assume a generative model for the data; instead, invariance is encouraged empirically by favoring feature subsets whose predictive behavior remains consistent across environments.

\noindent\underline{\emph{Differentiability.}}
Because the feature mask $s$ enters the predictors through element-wise input masking, the objective $U(s)$ is differentiable with respect to $s$ under mild assumptions on the predictor architectures.
Gradients $\nabla_s U(s)$ are computed via automatic differentiation through the frozen predictors and are stabilized using standard gradient clipping.
This differentiability enables gradient-based posterior inference in continuous mask space, which we exploit in the subsequent sampling stage.

\subsection{Guided Posterior Sampling}
\label{subsec:sampling}

Given the diffusion-based prior $p(s)$ and the stability-based likelihood $p(\mathcal{D}\mid s)$, we perform approximate posterior inference over feature selection masks using gradient-based sampling in continuous mask space. Direct sampling from the posterior
$p(s\mid\mathcal{D}) \propto p(\mathcal{D}\mid s)\,p(s)$
is intractable due to the high dimensionality of the space and the absence of a closed-form density. We therefore employ a Langevin dynamics-based sampling procedure that integrates information from both the learned prior and the stability-based likelihood.

\noindent\underline{\emph{Posterior score decomposition}}
The gradient of the log-posterior with respect to the mask $s$ decomposes as
\[
\nabla_s \log p(s \mid \mathcal{D})
=
\nabla_s \log p(s)
+
\nabla_s \log p(\mathcal{D} \mid s),
\]
where the first term corresponds to the score of the learned diffusion prior and the second term is given by the gradient of the stability-based objective defined in Section~\ref{subsec:likelihood}. This decomposition enables principled integration of structural plausibility and empirical performance-robustness signals during inference.

\noindent\underline{\emph{Annealed Langevin dynamics}}
Starting from an initial mask $s^{(0)}$ sampled uniformly from $[0,1]^p$, we iteratively update the mask according to
\begin{equation}
s^{(k+1)} =
s^{(k)}
+
\eta_k
\Big(
\nabla_s \log p(s^{(k)})
+
\beta\,\nabla_s \log p(\mathcal{D}\mid s^{(k)})
\Big)
+
\sqrt{2\eta_k}\,\epsilon^{(k)},
\label{eq:langevin}
\end{equation}
where $\{\eta_k\}$ is a decreasing step-size schedule, $\beta>0$ controls the relative influence of the likelihood term (i.e., likelihood tempering), and $\epsilon^{(k)} \sim \mathcal{N}(0, I)$ is Gaussian noise. After each update, the mask is projected onto $[0, 1]^p$ via clipping to ensure validity of continuous selection representation.

\noindent\underline{\emph{Score computation}}
The prior score $\nabla_s \log p(s)$ is obtained from the learned diffusion model by evaluating the score network at the final diffusion timestep. The likelihood gradient $\nabla_s \log p(\mathcal{D}\mid s)$ is computed via automatic differentiation through the frozen environment-specific predictors. To stabilize sampling, we normalize the prior and likelihood gradients before combination and apply standard gradient clipping.
To account for posterior uncertainty, we run multiple independent sampling chains in parallel and retain the final mask from each chain. The resulting set of samples provides an empirical approximation to the posterior distribution over feature subsets and supports downstream aggregation and discretization.

\subsection{Causal Guidance via Invariant Causal Prediction}
\label{subsec:icp}

While the stability-based likelihood in Section~\ref{subsec:likelihood} promotes empirical invariance through performance consistency across environments, it does not explicitly encode causal information. To further guide posterior inference toward stable feature subsets, we incorporate Invariant Causal Prediction (ICP) to bias posterior inference toward feature subsets that are more likely to satisfy causal invariance.

\noindent\underline{\emph{Invariant causal signals}}
Invariant Causal Prediction (ICP) identifies features whose conditional relationship with the target remains stable across environments under certain assumptions on the data-generating process.
Rather than treating ICP as a mechanism for causal discovery, we use it to obtain soft, continuous invariance scores that reflect the degree to which individual features are consistent with invariance across environments.

ICP is motivated by the principle that, for a set of causally relevant features $S$, the conditional distribution of the target should be invariant across environments, i.e.,
\[
Y \;\perp\!\!\!\perp\; E \;\mid\; X_S.
\]
In practice, testing this condition exactly is challenging in finite-sample and high-dimensional settings. We therefore use ICP-derived invariance scores only as weak guidance signals rather than as hard selection criteria.
We incorporate these invariance scores by biasing the mask pool construction used to train the diffusion prior. Specifically, features with higher invariance scores are sampled with higher probability when generating candidate masks. This increases the prior probability mass assigned to feature subsets that include such features, while preserving support over the entire mask space.
Importantly, this mechanism introduces no hard constraints: all feature subsets remain admissible under the prior, and the diffusion model learns a smooth distribution rather than enforcing deterministic inclusion or exclusion.

\noindent\underline{\emph{Interaction with posterior inference.}}
During guided posterior sampling (Section~\ref{subsec:sampling}), the stability-based likelihood continues to act as the primary arbiter of feature relevance.
As a result, feature subsets favored by ICP can be downweighted or rejected if they exhibit poor predictive performance or instability across environments.
This separation ensures that violations of ICP assumptions do not compromise the validity of the inference procedure.
By incorporating ICP as auxiliary causal guidance rather than as a hard constraint, CGDFS remains robust to imperfect or violated causal assumptions while still benefiting from invariance-based signals when they are informative. In practice, this guidance accelerates exploration of the feature subset space and improves sampling efficiency without altering the underlying inference objective.

\subsection{Discretization and Final Feature Selection}
\label{subsec:discretization}

The guided posterior sampling procedure produces a collection of continuous feature selection masks
$\{\tilde{s}^{(r)}\}_{r=1}^R \subset [0,1]^p$.
We convert these samples into discrete feature subsets in a post-hoc manner to obtain a final, interpretable selection.
For each posterior sample $\tilde{s}^{(r)}$, we derive a discrete candidate subset by selecting the indices of the top-$k$ entries,
\[
z^{(r)} = \mathrm{Top}\text{-}k(\tilde{s}^{(r)}),
\]
where $k$ is a user-specified target cardinality.
This maps continuous masks to feature subsets of fixed size while preserving relative feature importance within each sample.
Rather than selecting a single subset from a particular posterior sample, we aggregate information across samples by computing posterior inclusion frequencies
\[
\pi_j = \frac{1}{R} \sum_{r=1}^R \mathbb{I}\!\left(j \in z^{(r)}\right),
\]
which quantify how frequently feature $j$ appears among the top-$k$ selected features across posterior draws.
These frequencies provide an interpretable summary of feature relevance under the posterior distribution.

The final feature subset is obtained by selecting the $k$ features with the highest posterior inclusion frequencies.
This aggregation strategy leverages multiple posterior samples, reduces sensitivity to individual stochastic realizations, and yields a stable and interpretable feature selection.

\subsection{Theoretical Properties and Computational Considerations}
\label{subsec:theory}

\noindent\underline{\emph{Approximate Bayesian inference}}
CGDFS performs approximate posterior inference over feature selection masks by combining a learned diffusion prior with a stability-based likelihood and sampling via annealed Langevin dynamics.
The resulting dynamics target the posterior
\[
p(s \mid \mathcal{D}) \propto \exp\!\big( \log p(s) + \log p(\mathcal{D} \mid s) \big),
\]
but do not guarantee exact sampling due to the use of learned score models and non-convex objectives.
Accordingly, CGDFS emphasizes empirical posterior approximation and stable exploration of the feature subset space rather than asymptotic convergence guarantees.

\noindent\underline{\emph{Computational complexity}}
CGDFS consists of three computational components.
Learning the diffusion-based prior scales linearly with the size of the mask pool and the feature dimension.
Training environment-specific predictors scales linearly with the number of environments and data points per environment and can be parallelized.
Posterior sampling scales linearly with the number of sampling steps, chains, environments, and features.
Overall, the method scales polynomially in the number of features and environments, making it suitable for moderate-dimensional tabular datasets.

%% file: sections/experiments.tex

\section{Experiments}
\label{sec:experiments}

We evaluate CGDFS on a diverse set of tabular datasets under distribution shift, covering both classification and regression tasks. Our experiments are designed to answer the following research questions:
\textbf{RQ1.} Does CGDFS improve downstream predictive performance relative to competitive feature selection baselines?\\
\textbf{RQ2.} Does CGDFS produce more stable predictive behavior across heterogeneous environments?\\
\textbf{RQ3.} Which components of the proposed framework are critical to its empirical effectiveness?\\
\textbf{RQ4.} Does CGDFS yield interpretable and reproducible feature selections that reflect posterior uncertainty?

\vspace{-0.2cm}
\subsection{Experimental Setup}
\label{subsec:setup}

\subsubsection{Datasets and Tasks}
\label{subsec:datasets}
We evaluate CGDFS on a collection of 12 tabular datasets drawn from diverse application domains, sourced from the UCI Machine Learning Repository and OpenML. These datasets are selected to reflect a range of feature dimensionalities, sample sizes, and degrees of distributional heterogeneity induced through environment construction. We consider two widely studied predictive tasks in machine learning: classification (C) and regression (R).
Table~\ref{tab:overall_performance} summarizes the key characteristics of each dataset.

\subsubsection{Baselines}
\label{subsec:baselines}

We compare CGDFS against a diverse set of feature selection baselines that includes correlation-based, stability-aware, optimization-based, and invariance-based approaches. All baselines are evaluated under the same experimental protocol and use identical downstream models and feature budgets to ensure fair comparison.
\textbf{LASSO / Elastic Net:} Embedded linear feature selection methods that induce sparsity through $\ell_1$ or mixed $\ell_1/\ell_2$ regularization. Features are selected based on the magnitude of learned coefficients.
\textbf{Mutual Information (MI):} A univariate filter method that ranks features by estimated mutual information with the target variable and selects the top-$k$ features.
\textbf{Random Forest Importance:} Tree-based feature importance scores computed using Random Forests. Features are ranked by impurity-based or permutation importance, and the top-$k$ features are selected.
\textbf{Stability Selection:} A resampling-based wrapper method that aggregates feature selections across bootstrap samples to identify features that are consistently selected, with final selection based on selection frequency.
\textbf{Greedy Forward Selection:} A wrapper method that iteratively adds the feature that most improves validation performance, stopping when $k$ features have been selected.
\textbf{Gradient Ascent on Stability Objective:} A continuous optimization baseline that directly minimizes the stability-based objective defined in Section~\ref{subsec:likelihood} using gradient-based optimization over continuous feature masks, followed by top-$k$ discretization. This baseline isolates the effect of posterior sampling by replacing it with point-estimate optimization.
\textbf{Invariant Causal Prediction (ICP):} A causal feature selection baseline that selects features using invariance tests across environments under standard ICP assumptions, without posterior inference or stability-based optimization.
For all baselines, we evaluate the selected feature subsets using the same downstream predictive models (logistic regression for classification and ridge regression for regression). Hyperparameters for each baseline are tuned using validation environments only. Implementation details and hyperparameter grids are provided in the appendix.

\subsubsection{Environment Construction}
\label{subsec:env_construction}

To evaluate robustness under distribution shift, we explicitly construct multiple environments for each dataset by inducing controlled covariate shifts.
Specifically, environments are generated by partitioning the data according to predefined attributes or by applying environment-specific transformations to selected feature groups, resulting in distinct marginal feature distributions across environments while preserving the prediction task.
For each dataset, we construct $E$ environments, each corresponding to a different data distribution.
Environment splits are fixed across all methods and random seeds.
Crucially, environment labels are used only for evaluation of stability and for training environment-specific predictors; no environment information is provided to downstream models at test time.

\subsubsection{Evaluation Protocol and Metrics}
\label{subsec:metrics}

No test environment data is used during mask pool construction, diffusion prior training, hyperparameter tuning, or posterior sampling.

\noindent\underline{\emph{Downstream model and feature budget.}}
To evaluate each feature set produced by CGDFS and the baselines, we train a simple, standardized downstream model on the pooled training data (across training environments) using only the selected features:
For classification tasks, we use Logistic regression with $\ell_2$ regularization and for Regression, we use Ridge regression.
We report results for a fixed feature budget $k$ (reported in each table). Across all methods, the same $k$ is enforced to ensure a fair comparison. 

\noindent\underline{\emph{Metrics.}}
For Classification, we report F1-macro, and for Regression, we report MSE.
Other metrics are reported in the appendix.

\noindent\underline{\emph{Per-environment reporting and stability measures.}}
To evaluate robustness to distribution shift, we report both aggregate mean performance and environment-wise results. We also quantify stability by reporting the variance (and standard deviation) of the primary metric across test environments; lower variance indicates more stable predictive behavior under distributional heterogeneity.

\noindent\underline{\emph{Implementation and reproducibility.}}
All experiments are implemented in PyTorch and scikit-learn.
Random seeds, dataset preprocessing scripts, environment generation code, and hyperparameter grids are provided in the appendix. The code, experiments and data are available at https://anonymous.4open.science/r/CGD-FS-B01B/

\vspace{-0.2cm}
\subsection{Experimental Results}
\label{subsec:exp_results}

\subsubsection{Overall Predictive Performance}
\label{subsubsec:overall_results}

Table~\ref{tab:overall_performance} reports overall predictive performance across 12 datasets under distribution shift.
Across all six classification datasets, CGDFS achieves the highest F1-macro, outperforming correlation-based methods, stability-selection baselines, and direct optimization of the stability objective. Performance gains are observed consistently across datasets with varying dimensionality and sample size, indicating that posterior inference remains effective across diverse tabular settings. On the regression benchmarks, CGDFS attains the lowest mean squared error on all datasets. Compared to direct gradient-based optimization of the stability objective (Grad-Stab), CGDFS consistently reduces prediction error, highlighting the benefit of sampling-based posterior inference over point-estimate optimization under distribution shift. Invariant causal selection via ICP improves over purely correlation-driven baselines, but remains inferior to CGDFS, suggesting that invariance signals are most effective when integrated as soft guidance within a posterior framework.
Overall, these results indicate that CGDFS selects feature subsets that yield consistently stronger downstream performance across both classification and regression tasks.

\input{tables/overall_perf}

\vspace{-0.2cm}
\subsubsection{Robustness and Stability Analysis}
\label{subsubsec:stability}

\begin{figure*}[htbp]
\centering
\subfigure[UCI Credit]{
\includegraphics[width=4.3cm, height=3.5cm]{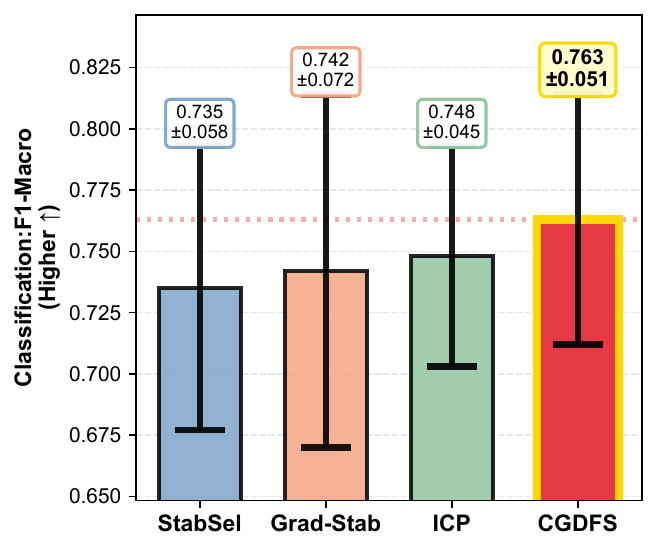}
}
\hspace{-3mm}
\subfigure[Activity]{ 
\includegraphics[width=4.3cm, height=3.5cm]{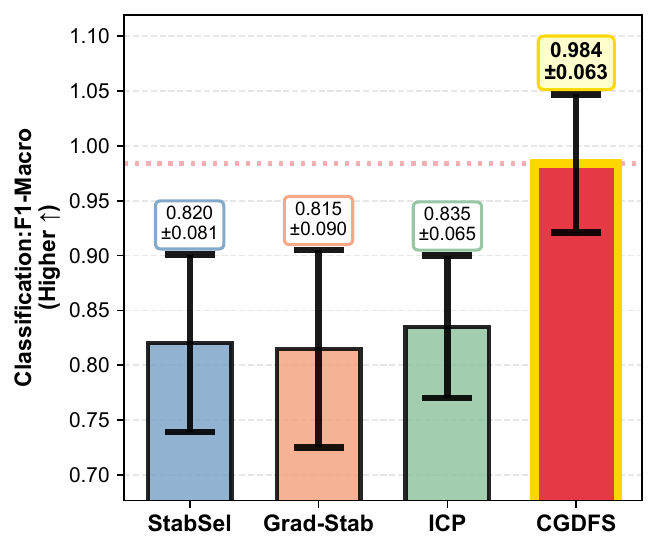}
}
\hspace{-3mm}
\subfigure[Openml-618]{
\includegraphics[width=4.3cm, height=3.5cm]{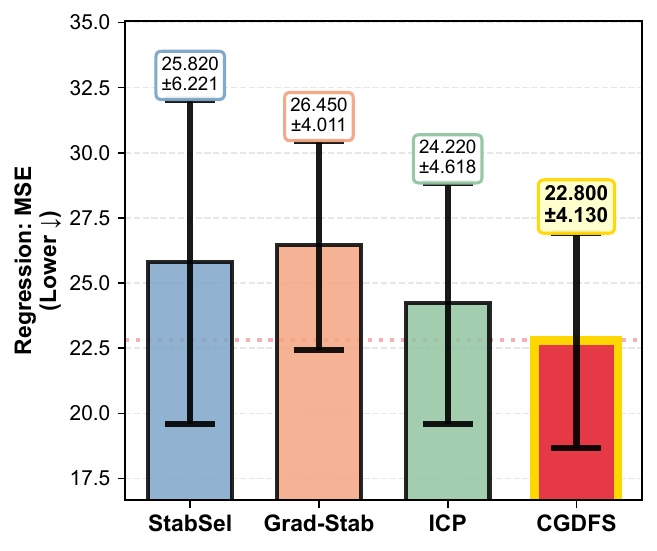}
}
\hspace{-3mm}
\subfigure[Openml-637]{ 
\includegraphics[width=4.3cm, height=3.5cm]{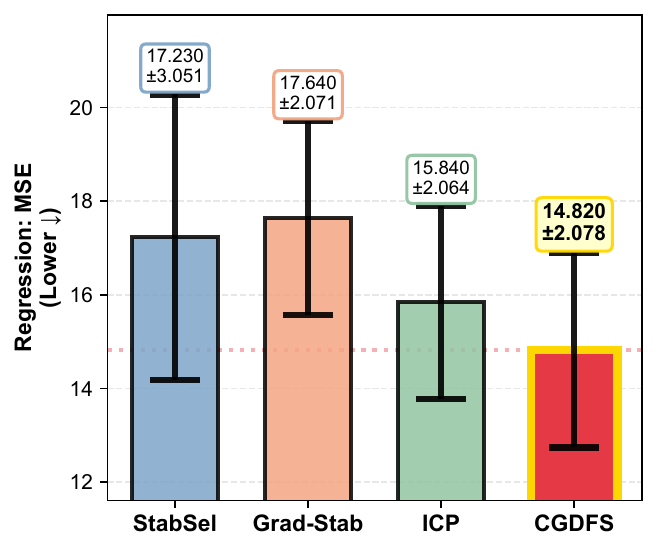}
}

\vspace{-0.25cm}
\caption{Cross-Environment Stability Analysis}
\label{fig:stability_study}
\vspace{-0.25cm}
\end{figure*}

Table~\ref{tab:stability_by_baseline} and Figure~\ref{fig:stability_study} report cross-environment performance (mean $\pm$ standard deviation) for stability-aware baselines and CGDFS.
Across all four datasets, CGDFS achieves strong average performance while maintaining competitive or improved stability relative to alternative methods. On the classification datasets, CGDFS attains the highest mean F1-macro.
For UCI Credit, CGDFS improves average performance over Stability Selection and Grad-Stab while exhibiting comparable variability across environments. On the Activity dataset, CGDFS substantially outperforms all baselines in average performance, while maintaining lower variance than Grad-Stab and Stability Selection. For regression tasks, CGDFS consistently achieves the lowest mean squared error.
On OpenML-618, CGDFS exhibits variability comparable to competing methods, while on OpenML-637 it attains both the lowest mean error and the lowest cross-environment variance. Although ICP often reduces variability, this reduction is frequently accompanied by degraded average performance. In contrast, CGDFS achieves a favorable balance between predictive accuracy and cross-environment stability, indicating robustness under distribution shift.

\vspace{-0.2cm}
\subsubsection{Ablation Study}
\label{subsec:ablation}

\begin{figure}[htbp]
\centering
\subfigure[UCI Credit]{
\includegraphics[width=4cm]{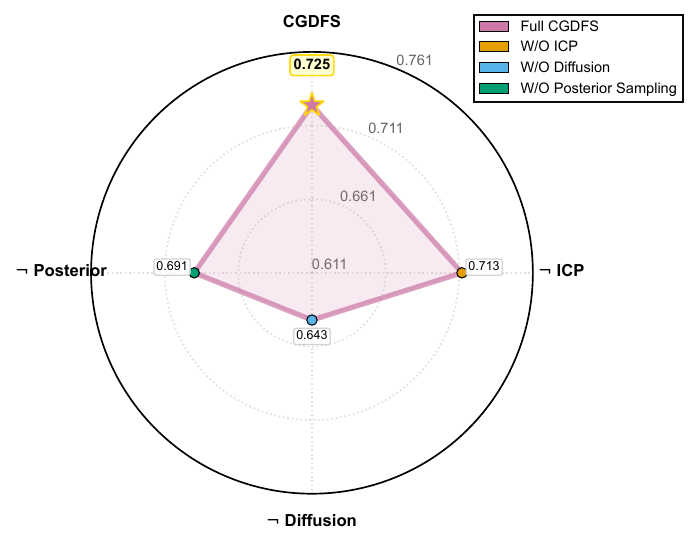}
}
\hspace{-3mm}
\subfigure[Pima Indian]{ 
\includegraphics[width=4cm]{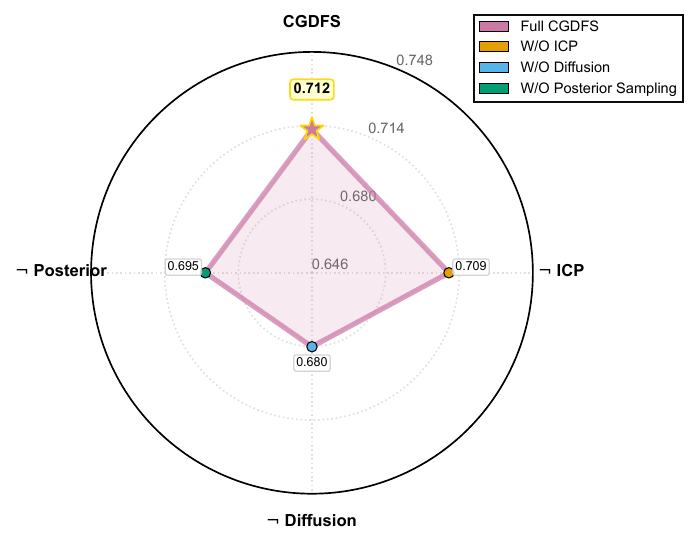}
}
\hspace{-3mm}
\subfigure[Openml-618]{
\includegraphics[width=4cm]{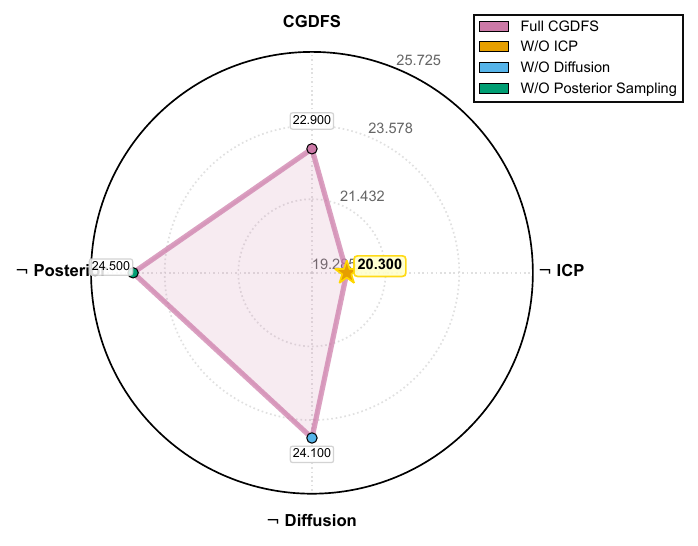}
}
\hspace{-3mm}
\subfigure[Boston Housing]{ 
\includegraphics[width=4cm, height=3cm]{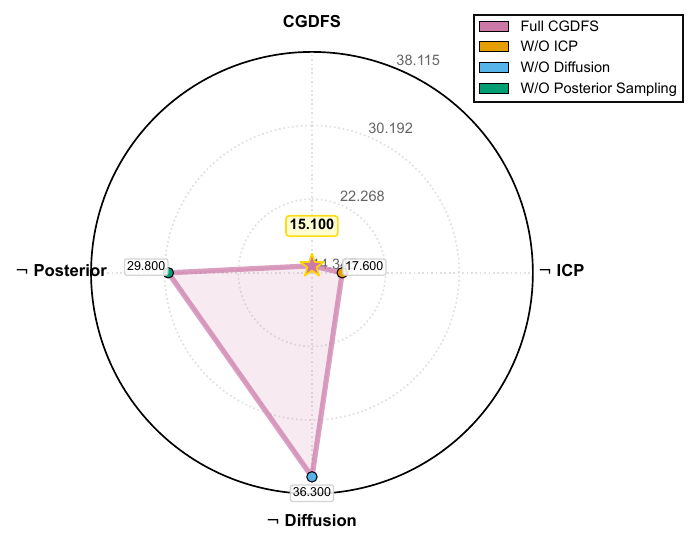}
}
\vspace{-0.25cm}
\caption{CGDFS - Ablation Study}
\label{fig:ablation_study}
\vspace{-0.25cm}
\end{figure}

We analyze the contribution of individual components of CGDFS by systematically removing the diffusion prior, posterior sampling, and causal guidance as seen in figure~\ref{fig:ablation_study}.

\noindent\underline{\emph{Effect of the Diffusion Prior.}}
Removing the diffusion-based prior leads to substantial degradation in predictive performance across all datasets on both classification and regression tasks. While stability variance remains comparable in some cases, the resulting feature subsets are consistently less predictive, indicating that posterior sampling alone is insufficient without a structured prior. This confirms that the diffusion prior is critical for guiding exploration toward meaningful regions of the combinatorial feature space.

\noindent\underline{\emph{Posterior Sampling vs. Optimization.}}
Replacing posterior sampling with direct gradient-based optimization of the stability objective (Grad-Stab) results in lower predictive performance and higher cross-environment variance. Though Grad-Stab yields feature subsets with competitive predictive performance, it shows increased sensitivity to initialization and unstable behavior across environments. In contrast, posterior sampling enables CGDFS to aggregate information from multiple high-probability solutions, yielding more stable results.

\noindent\underline{\emph{Effect of Causal Guidance (ICP).}}
Incorporating invariant causal prediction as soft guidance improves cross-environment stability on both classification and regression datasets. Notably, CGDFS without ICP remains competitive, while ICP alone performs worse than all posterior-based variants. These results demonstrate that causal guidance is beneficial but not essential, and is most effective when integrated as a soft prior within the posterior inference framework rather than enforced as a hard selection criterion.

\subsubsection{Interpretability and Uncertainty Analysis}
\label{subsec:interpretability}

We evaluate the interpretability of the selected feature subsets by examining the posterior inclusion frequencies and feature stability across runs.

\textbf{Posterior Inclusion Frequencies: }CGDFS yields a posterior distribution over the feature subsets, which is represented using posterior inclusion frequency. This metric quantifies how frequently each feature appears among the top-$k$ selected features across posterior samples. This provides an interpretable measure of feature relevance that directly reflects uncertainty. Across datasets, the selected features exhibit higher inclusion frequencies, while low-importance features exhibit a low posterior inclusion rate. This enables interpretable ranking of features and avoids reliance on a single deterministic selection.
\textbf{Feature Stability Across Runs:}We also evaluate the overlap of selected feature subsets across independent runs with different random seeds. CGDFS consistently produces higher overlap compared to correlation and optimization-based baselines, indicating reduced sensitivity to initialization and sampling noise. It reasons to the posterior aggregation in our framework, which mitigates variance and outputs more reliable feature subsets. These analyses show that our framework CGDFS yields feature subsets that are interpretable, with uncertainty explicitly quantified through posterior runs.

\input{tables/uncertainty}

\subsubsection{Computational Complexity and Runtime}
\label{subsec:runtime}
We evaluate the computational cost of the three stages of CGDFS: diffusion prior training, environment-specific predictor training, and guided posterior sampling. Overall, computational complexity scales polynomially with the number of features and environments. Environment-specific computations are fully parallelizable. In practice, runtime is dominated by diffusion prior training, while posterior sampling scales linearly with the number of features and environments. Across all datasets, CGDFS completes feature selection within a practical time budget and exhibits runtime comparable to stability-based and invariance-based baselines that require repeated model training.

%% file: tables/overall_perf.tex
\begin{table*}[t]
\centering
\small
\setlength{\tabcolsep}{4pt}
\caption{Overall predictive performance across 12 datasets under distribution shift.
For classification tasks, we report F1-macro (higher is better).
For regression tasks, we report MSE (lower is better).
All results are averaged over multiple random seeds.
}
\vspace{-0.2cm}
\label{tab:overall_performance}
\begin{tabular}{@{}lcc|cccccccc@{}}
\toprule
\multirow{2}{*}{\textbf{Dataset}} 
& \multirow{2}{*}{\textbf{\#Samples}} 
& \multirow{2}{*}{\textbf{\#Features}} 
& \multicolumn{8}{c}{\textbf{Feature Selection Method}} \\
\cmidrule(lr){4-11}
& & 
& LASSO & MI & RF & StabSel & Greedy & Grad-Stab & ICP & \textbf{CGDFS} \\
\midrule
\multicolumn{11}{c}{\textbf{Classification (F1-Macro $\uparrow$)}} \\
\midrule
German Credit     & 1001  & 24  & 0.680 & 0.665 & 0.695 & 0.692 & 0.705 & 0.688 & 0.712 & \textbf{0.725} \\
Ionosphere        & 351      & 34  & 0.825 & 0.810 & 0.838 & 0.835 & 0.842 & 0.830 & 0.848 & \textbf{0.985} \\
Pima Indian       & 768      & 9   & 0.655 & 0.642 & 0.668 & 0.665 & 0.672 & 0.660 & 0.680 & \textbf{0.712} \\
Spam Base         & 4601  & 58  & 0.885 & 0.878 & 0.892 & 0.890 & 0.896 & 0.888 & 0.902 & \textbf{0.915} \\
UCI Credit Card   & 30000 & 25  & 0.725 & 0.715 & 0.738 & 0.735 & 0.742 & 0.730 & 0.748 & \textbf{0.762} \\
Activity          & 10{,}299 & 561 & 0.812 & 0.805 & 0.825 & 0.820 & 0.828 & 0.815 & 0.835 & \textbf{0.984} \\
\midrule
\multicolumn{11}{c}{\textbf{Regression (MSE $\downarrow$)}} \\
\midrule
Boston Housing    & 506      & 13  & 18.06 & 19.82 & 17.23 & 17.48 & 16.81 & 17.81 & 16.16 & \textbf{15.06} \\
OpenML-586        & 1000  & 25  & 12.40 & 13.54 & 11.90 & 12.11 & 11.70 & 12.25 & 11.22 & \textbf{10.37} \\
OpenML-589        & 1000  & 25  & 8.70  & 9.49  & 8.29  & 8.53  & 8.12  & 8.64  & 7.73  & \textbf{7.18} \\
OpenML-637        & 1000  & 50  & 17.47 & 18.92 & 16.81 & 17.23 & 16.63 & 17.64 & 15.84 & \textbf{14.82} \\
OpenML-616        & 500      & 50  & 13.54 & 14.59 & 13.11 & 13.32 & 12.82 & 13.69 & 12.39 & \textbf{11.70} \\
OpenML-618        & 1000  & 50  & 26.21 & 27.88 & 25.53 & 25.82 & 25.21 & 26.45 & 24.22 & \textbf{22.85} \\
\bottomrule
\end{tabular}
\end{table*}

%% file: tables/uncertainty.tex

\begin{table}[t]
\centering
\footnotesize
\setlength{\tabcolsep}{3pt}
\caption{Uncertainty statistics of feature selection for CGDFS.
Uncertainty is computed from posterior inclusion frequencies, where higher values indicate greater feature uncertainty.}
\label{tab:uncertainty_analysis}
\begin{threeparttable}
\begin{tabular}{@{}lrrr@{}}
\toprule
\textbf{Uncertainty Metric} & \textbf{Activity} & \textbf{Spam Base} & \textbf{Boston Housing} \\
\midrule
\multicolumn{4}{c}{\textit{Central Tendency}} \\
\midrule
Mean Uncertainty & 0.1847 & 0.0832 & 0.2761 \\
Median Uncertainty & 0.1693 & 0.0654 & 0.2453 \\
Std.\ Dev.\ Uncertainty & 0.1214 & 0.1456 & 0.2422 \\
\midrule
\multicolumn{4}{c}{\textit{Extreme Values}} \\
\midrule
Minimum Uncertainty & 0.0008 & 0.0001 & 0.0001 \\
Maximum Uncertainty & 0.4998 & 0.4856 & 0.4999 \\
\midrule
\multicolumn{4}{c}{\textit{Feature Uncertainty Distribution}} \\
\midrule
Q1 (0--0.25 uncertainty) & 72.25\% & 88.00\% & 52.00\% \\
Q2 (0.25--0.50 uncertainty) & 27.75\% & 12.00\% & 48.00\% \\
Q3 ($>$0.50 uncertainty) & 0.00\% & 0.00\% & 0.00\% \\
\midrule
\multicolumn{4}{c}{\textit{Interpretation}} \\
\midrule
Mean Confidence ($1-\text{Uncertainty}$) & 81.53\% & 91.68\% & 59.26\% \\
\bottomrule

\vspace{-0.5cm}
\end{tabular}
\end{threeparttable}
\end{table}

%% file: sections/related_works.tex

\section{Related Work}
\label{sec:related_work}

\subsection{Feature Selection under Distribution Shift}
\label{subsec:rw_robust_fs}

Feature selection has traditionally been studied to improve generalization and interpretability, with classical approaches broadly categorized as filter, wrapper, and embedded methods~\cite{yu2003feature, hoque2014mifs, el2016review, zou2005regularization, muthukrishnan2016lasso, ying2025survey}. While effective in i.i.d.\ settings, these methods typically optimize predictive performance under a single training distribution and are therefore vulnerable to spurious correlations that fail under distribution shift. To improve robustness, stability-based feature selection methods emphasize consistency across resampling or data perturbations~\cite{meinshausen2010stability, shah2013variable, buyukkecceci2023comprehensive}. However, these approaches rely on heuristic aggregation and do not explicitly model distributional heterogeneity or uncertainty over feature subsets. Related work on domain adaptation and covariate shift~\cite{gong2016domain, li2018deep, ruder2017overview} primarily focuses on representation learning rather than explicit feature subset selection. In contrast, CGDFS directly addresses feature selection under distribution shift by explicitly modeling predictive performance and stability across environments within a unified probabilistic framework.

\subsection{Invariant Learning and Causal Feature Selection}
\label{subsec:rw_causal}

Invariant and causal learning methods aim to improve robustness by exploiting the principle that causal relationships remain stable across environments, whereas spurious correlations do not~\cite{pearl2009causal, peters2017elements, buhlmann2020invariance, malarkkan2024multi, malarkkan2025causal, malarkkan2025incremental, arun2025delta, malarkkan2026causally}. Approaches such as Invariant Risk Minimization (IRM)~\cite{arjovsky2019invariant, ahuja2020invariant} and Invariant Causal Prediction (ICP)~\cite{peters2016causal, heinze2018invariant} operationalize this idea by enforcing or testing invariance across environments. Several works have explored causal feature selection using invariance-based criteria and regularization~\cite{magliacane2018domain, rojas2018invariant, arun2025delta}. However, these methods often rely on hard constraints, combinatorial subset testing, or strong assumptions about environment semantics, which limits scalability and robustness in high-dimensional settings. Moreover, they typically return a single accepted feature set and do not quantify uncertainty. CGDFS differs by treating invariance signals as soft causal guidance and integrating them into posterior inference over feature subsets, allowing empirical performance and stability to override imperfect causal assumptions.

\subsection{Probabilistic Inference and Diffusion Models}
\label{subsec:rw_diffusion}

Bayesian formulations of feature selection and model uncertainty date back to Bayesian variable selection and model averaging~\cite{george1993variable, mitchell1988bayesian}. Sampling-based inference methods, including Markov chain Monte Carlo and Langevin dynamics~\cite{neal2011mcmc, welling2011bayesian, vollmer2015non}, provide principled tools for approximate posterior inference but are rarely used in modern feature selection pipelines due to scalability challenges. Recent work has explored feature selection via deep generative models~\cite{ying2024feature, gong2025neuro}, where subsets are generated sequentially using autoregressive or constrained generative processes. In parallel, score-based diffusion models have emerged as powerful tools for learning complex high-dimensional distributions~\cite{song2019generative, ho2020denoising, song2020score}, with guided diffusion techniques applied to structured generation and inverse problems~\cite{dhariwal2021diffusion, aali2023solving}. However, their application to feature selection remains largely unexplored. CGDFS bridges these lines of work by using diffusion models to learn a structured prior over feature subsets and combining it with stability-based likelihoods through guided Langevin dynamics, enabling scalable and uncertainty-aware feature selection under distribution shift.

%% file: sections/conclusion.tex

\section{Conclusion}

We studied feature selection under distribution shift and proposed \emph{Causally-Guided Diffusion for Stable Feature Selection} (CGDFS), a framework that formulates feature selection as approximate posterior inference over feature subsets. CGDFS jointly accounts for predictive performance and cross-environment stability by combining a diffusion-based prior over feature subsets with a stability-based likelihood defined across multiple environments.
Our approach departs from deterministic optimization by explicitly reasoning over a distribution of feature subsets. Posterior sampling enables exploration of multiple high-probability solutions in the combinatorial feature space, while posterior aggregation yields stable and reproducible selections. Invariance-based causal signals are incorporated as soft guidance through the prior, allowing empirical stability and predictive performance to remain the dominant criteria.
Empirical results on classification and regression tasks under controlled distribution shifts demonstrate that CGDFS consistently improves out-of-distribution performance and reduces performance variance across environments relative to strong feature selection baselines, including stability-based, optimization-based, and invariance-based methods. These results suggest that posterior inference provides a principled and effective alternative to point-estimate feature selection in non-i.i.d.\ settings.

%% file: sections/appendix.tex

\onecolumn
\appendix

\section{Theoretical Justification}
\label{app:theory}

This appendix provides concise theoretical justification for the design choices underlying CGDFS. The results below are stated under standard regularity assumptions and are intended to clarify the behavior of the energy-based posterior, the learned diffusion prior, and the final aggregation procedure. The statements are informal and asymptotic in nature, and are included to aid interpretation rather than to establish formal guarantees.

\subsection{Setup and Assumptions}

Let $s \in [0,1]^p$ denote a continuous feature selection mask and recall the stability-based objective
\[
\mathcal{J}(s)
=
\bar{\ell}(s)
+
\lambda_{\mathrm{var}}\,\mathrm{Var}_e\!\big(\ell_e(s)\big),
\]
with the induced energy-based posterior distribution
\[
p(s \mid \mathcal{D}) \propto \exp\!\big(-\mathcal{J}(s)\big)\,p(s),
\]
where $p(s)$ denotes the diffusion-learned prior. This distribution is interpreted as an energy-based posterior rather than as a likelihood-derived Bayesian posterior.

We make the following assumptions:
\begin{itemize}
    \item \textbf{A1 (Smoothness).} For each environment $e$, the loss $\ell_e(s)$ is continuously differentiable and $L$-Lipschitz on $[0,1]^p$.
    \item \textbf{A2 (Boundedness).} Losses are uniformly bounded: $\ell_e(s) \in [0, L_{\max}]$.
    \item \textbf{A3 (Prior support).} The prior density $p(s)$ is continuous and strictly positive on neighborhoods of masks of interest.
    \item \textbf{A4 (Score approximation).} Given sufficient mask-pool coverage and model capacity, the denoising score network provides a consistent approximation to the score of the perturbed mask distribution.
\end{itemize}

These assumptions are standard in score-based generative modeling and stability-based learning.

\subsection{Posterior Preference for Stable Feature Subsets}

\paragraph{Proposition 1.}
Under Assumptions A1--A3, for any two masks $s_a, s_b \in [0,1]^p$,
\[
\mathcal{J}(s_a) < \mathcal{J}(s_b)
\quad \Rightarrow \quad
p(s_a \mid \mathcal{D}) > p(s_b \mid \mathcal{D}),
\]
up to a multiplicative factor given by the prior ratio $p(s_a)/p(s_b)$.

\paragraph{Proof sketch.}
By definition,
\[
\frac{p(s_a \mid \mathcal{D})}{p(s_b \mid \mathcal{D})}
=
\frac{p(s_a)}{p(s_b)}
\exp\!\big(\mathcal{J}(s_b) - \mathcal{J}(s_a)\big).
\]
Assumption A3 ensures $p(s_a), p(s_b) > 0$. Since $\mathcal{J}(s_b) - \mathcal{J}(s_a) > 0$, the ratio exceeds one. \hfill $\square$

\paragraph{Implication.}
The posterior explicitly favors feature subsets that achieve better trade-offs between average predictive performance and cross-environment stability, justifying the likelihood construction used in CGDFS.

\subsection{Consistency of the Learned Diffusion Prior}

\paragraph{Proposition 2 (Consistency intuition).}
Under Assumptions A3--A4 and standard conditions on score-based diffusion training, the learned score network $s_\theta(s,t)$ approximates the score of the perturbed mask distribution. Evaluating the score network at the lowest-noise diffusion timestep yields an approximate gradient of $\log p(s)$ up to modeling and optimization error.

\paragraph{Proof sketch.}
Denoising score matching minimizes the expected squared error between the network output and injected Gaussian noise. Under sufficient model capacity and training accuracy, this objective is minimized by the true denoising function, which is algebraically related to the score of the perturbed density. As the noise variance decreases, this score approaches the gradient of the log-density of the underlying mask distribution. \hfill $\square$

\paragraph{Implication.}
This result supports using the learned diffusion score as a prior gradient during guided posterior sampling.

\subsection{Consistency of Posterior Inclusion Frequencies}

\paragraph{Proposition 3.}
Let $\{\tilde{s}^{(r)}\}_{r=1}^R$ denote approximate posterior samples obtained via guided Langevin dynamics, and let
$z^{(r)} = \mathrm{Top}\text{-}k(\tilde{s}^{(r)})$.
Define the empirical inclusion frequency
\[
\pi_j^{(R)} = \frac{1}{R}\sum_{r=1}^R \mathbb{I}\!\left(j \in z^{(r)}\right).
\]
Assuming the sampling procedure yields asymptotically unbiased samples from $p(s \mid \mathcal{D})$, then
\[
\pi_j^{(R)} \xrightarrow{\text{a.s.}}
\mathbb{P}_{p(s \mid \mathcal{D})}\!\left(j \in \mathrm{Top}\text{-}k(s)\right)
\quad \text{as } R \to \infty.
\]

\paragraph{Proof sketch.}
For fixed $j$, the indicator $\mathbb{I}(j \in \mathrm{Top}\text{-}k(s))$ is a bounded measurable function. By the strong law of large numbers, empirical averages converge almost surely to their expectation under the sampling distribution. \hfill $\square$

\paragraph{Implication.}
Posterior inclusion frequencies provide consistent estimates of feature relevance probabilities, and selecting features with the highest $\pi_j$ yields a stable decision rule under a fixed cardinality constraint.

\section{Hyperparameter Sensitivity}
\label{app:hyperparam}

We evaluate CGDFS sensitivity to key hyperparameters that most strongly affect inference: the variance penalty $\lambda_{\mathrm{var}}$ in the stability objective, the guidance weight $\beta$ in the Langevin sampler, and the sampling budget (number of chains $R$ and steps $K$). 
Results show that CGDFS performance and stability vary smoothly with these parameters, and that the chosen defaults lie in a stable region of the parameter space. Specifically, small $\lambda_{\mathrm{var}}$ values prioritize average performance while large values overemphasize stability; $\lambda_{\mathrm{var}}=0.1$ balances this trade-off. Moderate guidance ($\beta \approx 0.5$) effectively integrates the diffusion prior with the stability objective. Increasing $R$ and $K$ yields diminishing returns, with $R=10$ and $K=100$ providing a practical accuracy--efficiency trade-off.

\begin{table}[t]
\centering
\footnotesize
\setlength{\tabcolsep}{4pt}
\caption{Hyperparameter defaults and sweep ranges used in our experiments.}
\label{tab:hyper_defaults}
\begin{tabular}{lcc}
\toprule
Hyperparameter & Default & Sweep range \\
\midrule
$\lambda_{\mathrm{var}}$ & 0.1 & \{0.0, 0.01, 0.05, 0.1, 0.2, 0.5\} \\
Guidance $\beta$ & 0.5 & \{0.0, 0.25, 0.5, 0.75, 1.0\} \\
Langevin steps $K$ & 100 & \{20, 50, 100, 200\} \\
Chains $R$ & 10 & \{1, 3, 5, 10\} \\
Mask pool size $M$ & 500 & \{100, 300, 500\} \\
Diffusion steps $T$ & 100 & \{50, 100\} \\
Prior learning rate & $1\times10^{-4}$ & \{1e{-5}, 1e{-4}, 5e{-4}\} \\
Predictor hidden dim & 128 & \{64, 128, 256\} \\
\bottomrule
\end{tabular}
\end{table}

\section{Limitations}
\label{app:limitations}

CGDFS has several limitations that point to directions for future work. 1) Posterior inference is approximate: guided Langevin dynamics with a learned diffusion prior does not guarantee exact sampling from the target distribution, and performance depends on step-size schedules, noise levels, and prior accuracy. 2) The stability objective relies on a finite set of environments; if environments fail to expose relevant distribution shifts, stability estimates may be optimistic. 3) The learned diffusion prior reflects the quality and diversity of the mask pool; poorly constructed pools may bias exploration despite the posterior correction. 4) While ICP-based guidance can accelerate exploration, it depends on assumptions that may be violated in practice; CGDFS mitigates this risk by treating ICP as optional soft guidance rather than a constraint, but incorrect invariance signals may still affect sampling efficiency. Finally, the current implementation targets moderate-dimensional tabular data; scaling to very high-dimensional feature spaces remains an open challenge.

\section{Additional Experimental Results}
\label{app:extra_results}

\input{tables/ablation}
\input{tables/stability}

%% file: tables/ablation.tex

\begin{table*}[t]
\centering
\caption{Ablation and stability analysis of CGDFS
For classification, we report F1-macro (higher is better).
For regression, we report MSE (lower is better).
Stability is measured as the standard deviation of the primary metric across test environments (lower is better).
}
\label{tab:ablation_stability}
\setlength{\tabcolsep}{3.5pt}
\begin{tabular}{lcc|cc|cc|cc}
\toprule
\multirow{2}{*}{Method} 
& \multicolumn{2}{c}{UCI Credit (C)} 
& \multicolumn{2}{c}{Pima Indian (C)} 
& \multicolumn{2}{c}{Boston Housing (R)} 
& \multicolumn{2}{c}{OpenML-618 (R)} \\
\cmidrule(lr){2-3}\cmidrule(lr){4-5}\cmidrule(lr){6-7}\cmidrule(lr){8-9}
& F1 $\uparrow$ & Std $\downarrow$
& F1 $\uparrow$ & Std $\downarrow$
& MSE $\downarrow$ & Std $\downarrow$
& MSE $\downarrow$ & Std $\downarrow$ \\
\midrule
Full CGDFSS 
& \textbf{0.725} & \textbf{0.051}
& \textbf{0.712} & \textbf{0.034}
& \textbf{15.1} & \textbf{1.60}
& 22.9 & \textbf{1.13} \\

W/O ICP 
& 0.713 & 0.088
& 0.709 & 0.055
& 17.6 & 2.10
& \textbf{20.3} & 1.73 \\

W/O Diffusion 
& 0.643 & 0.058
& 0.680 & 0.039
& 36.3 & 1.67
& 24.1 & 1.42 \\

W/O Posterior Sampling
& 0.691 & 0.079
& 0.695 & 0.061
& 29.8 & 3.14
& 24.5 & 1.45 \\

\bottomrule
\end{tabular}
\end{table*}

%% file: tables/stability.tex


\begin{table}[t]
\centering
\small
\setlength{\tabcolsep}{10pt}
\caption{Cross-environment stability (mean $\pm$ standard deviation across test environments).
For classification datasets (UCI Credit, Activity), we report F1-macro (higher is better).
For regression datasets (OpenML-618, OpenML-637), we report MSE (lower is better).}
\label{tab:stability_by_baseline}
\begin{tabular}{lcccc}
\toprule
\textbf{Dataset} 
& \textbf{Stability Sel.} 
& \textbf{Grad-Stab} 
& \textbf{ICP} 
& \textbf{CGDFS} \\
\midrule
UCI Credit (C)  
& 0.735 $\pm$ 0.058 
& 0.742 $\pm$ 0.072 
& 0.748 $\pm$ 0.045 
& \textbf{0.763 $\pm$ 0.051} \\

Activity (C)    
& 0.820 $\pm$ 0.081 
& 0.815 $\pm$ 0.090 
& 0.835 $\pm$ 0.065 
& \textbf{0.984 $\pm$ 0.063} \\

\midrule
OpenML-618 (R)  
& 25.82 $\pm$ 6.22 
& 26.45 $\pm$ 4.01 
& 24.22 $\pm$ 4.62 
& \textbf{22.85 $\pm$ 4.13} \\

OpenML-637 (R)  
& 17.23 $\pm$ 3.05 
& 17.64 $\pm$ 2.07 
& 15.84 $\pm$ 2.06 
& \textbf{14.82 $\pm$ 2.08} \\
\bottomrule
\end{tabular}
\end{table}

%% file: main.bib
@inproceedings{yu2003feature,
  title={Feature selection for high-dimensional data: A fast correlation-based filter solution},
  author={Yu, Lei and Liu, Huan},
  booktitle={Proceedings of the 20th international conference on machine learning (ICML-03)},
  pages={856--863},
  year={2003}
}

@article{hoque2014mifs,
  title={MIFS-ND: A mutual information-based feature selection method},
  author={Hoque, Nazrul and Bhattacharyya, Dhruba K and Kalita, Jugal K},
  journal={Expert systems with applications},
  volume={41},
  number={14},
  pages={6371--6385},
  year={2014},
  publisher={Elsevier}
}

@inproceedings{el2016review,
  title={Review on wrapper feature selection approaches},
  author={El Aboudi, Naoual and Benhlima, Laila},
  booktitle={2016 international conference on engineering \& MIS (ICEMIS)},
  pages={1--5},
  year={2016},
  organization={IEEE}
}

@inproceedings{muthukrishnan2016lasso,
  title={LASSO: A feature selection technique in predictive modeling for machine learning},
  author={Muthukrishnan, Ramakrishnan and Rohini, RLASSO},
  booktitle={2016 IEEE international conference on advances in computer applications (ICACA)},
  pages={18--20},
  year={2016},
  organization={Ieee}
}

@article{zou2005regularization,
  title={Regularization and variable selection via the elastic net},
  author={Zou, Hui and Hastie, Trevor},
  journal={Journal of the Royal Statistical Society Series B: Statistical Methodology},
  volume={67},
  number={2},
  pages={301--320},
  year={2005},
  publisher={Oxford University Press}
}

@article{peters2016causal,
  title={Causal inference by using invariant prediction: identification and confidence intervals},
  author={Peters, Jonas and B{\"u}hlmann, Peter and Meinshausen, Nicolai},
  journal={Journal of the Royal Statistical Society Series B: Statistical Methodology},
  volume={78},
  number={5},
  pages={947--1012},
  year={2016},
  publisher={Oxford University Press}
}

@article{meinshausen2010stability,
  title={Stability selection},
  author={Meinshausen, Nicolai and B{\"u}hlmann, Peter},
  journal={Journal of the Royal Statistical Society Series B: Statistical Methodology},
  volume={72},
  number={4},
  pages={417--473},
  year={2010},
  publisher={Oxford University Press}
}

@article{shah2013variable,
  title={Variable selection with error control: another look at stability selection},
  author={Shah, Rajen D and Samworth, Richard J},
  journal={Journal of the Royal Statistical Society Series B: Statistical Methodology},
  volume={75},
  number={1},
  pages={55--80},
  year={2013},
  publisher={Oxford University Press}
}

@article{buyukkecceci2023comprehensive,
  title={A comprehensive review of feature selection and feature selection stability in machine learning},
  author={B{\"u}y{\"u}kke{\c{c}}eci, Mustafa and Okur, Mehmet Cudi},
  journal={Gazi University Journal of Science},
  volume={36},
  number={4},
  pages={1506--1520},
  year={2023},
  publisher={Gazi University}
}

@inproceedings{gong2016domain,
  title={Domain adaptation with conditional transferable components},
  author={Gong, Mingming and Zhang, Kun and Liu, Tongliang and Tao, Dacheng and Glymour, Clark and Sch{\"o}lkopf, Bernhard},
  booktitle={International conference on machine learning},
  pages={2839--2848},
  year={2016},
  organization={PMLR}
}

@inproceedings{li2018deep,
  title={Deep domain generalization via conditional invariant adversarial networks},
  author={Li, Ya and Tian, Xinmei and Gong, Mingming and Liu, Yajing and Liu, Tongliang and Zhang, Kun and Tao, Dacheng},
  booktitle={Proceedings of the European conference on computer vision (ECCV)},
  pages={624--639},
  year={2018}
}

@article{ruder2017overview,
  title={An overview of multi-task learning in deep neural networks},
  author={Ruder, Sebastian},
  journal={arXiv preprint arXiv:1706.05098},
  year={2017}
}

@article{pearl2009causal,
  title={Causal inference in statistics: An overview},
  author={Pearl, Judea},
  year={2009}
}

@book{peters2017elements,
  title={Elements of causal inference: foundations and learning algorithms},
  author={Peters, Jonas and Janzing, Dominik and Sch{\"o}lkopf, Bernhard},
  year={2017},
  publisher={The MIT press}
}

@article{buhlmann2020invariance,
  title={Invariance, causality and robustness},
  author={B{\"u}hlmann, Peter},
  journal={Statistical Science},
  volume={35},
  number={3},
  pages={404--426},
  year={2020},
  publisher={JSTOR}
}

@article{arjovsky2019invariant,
  title={Invariant risk minimization},
  author={Arjovsky, Martin and Bottou, L{\'e}on and Gulrajani, Ishaan and Lopez-Paz, David},
  journal={arXiv preprint arXiv:1907.02893},
  year={2019}
}

@inproceedings{ahuja2020invariant,
  title={Invariant risk minimization games},
  author={Ahuja, Kartik and Shanmugam, Karthikeyan and Varshney, Kush and Dhurandhar, Amit},
  booktitle={International Conference on Machine Learning},
  pages={145--155},
  year={2020},
  organization={PMLR}
}

@article{heinze2018invariant,
  title={Invariant causal prediction for nonlinear models},
  author={Heinze-Deml, Christina and Peters, Jonas and Meinshausen, Nicolai},
  journal={Journal of Causal Inference},
  volume={6},
  number={2},
  pages={20170016},
  year={2018},
  publisher={De Gruyter}
}

@article{magliacane2018domain,
  title={Domain adaptation by using causal inference to predict invariant conditional distributions},
  author={Magliacane, Sara and Van Ommen, Thijs and Claassen, Tom and Bongers, Stephan and Versteeg, Philip and Mooij, Joris M},
  journal={Advances in neural information processing systems},
  volume={31},
  year={2018}
}

@article{rojas2018invariant,
  title={Invariant models for causal transfer learning},
  author={Rojas-Carulla, Mateo and Sch{\"o}lkopf, Bernhard and Turner, Richard and Peters, Jonas},
  journal={Journal of Machine Learning Research},
  volume={19},
  number={36},
  pages={1--34},
  year={2018}
}

@article{george1993variable,
  title={Variable selection via Gibbs sampling},
  author={George, Edward I and McCulloch, Robert E},
  journal={Journal of the American Statistical Association},
  volume={88},
  number={423},
  pages={881--889},
  year={1993},
  publisher={Taylor \& Francis}
}

@article{mitchell1988bayesian,
  title={Bayesian variable selection in linear regression},
  author={Mitchell, Toby J and Beauchamp, John J},
  journal={Journal of the american statistical association},
  volume={83},
  number={404},
  pages={1023--1032},
  year={1988},
  publisher={Taylor \& Francis}
}

@article{neal2011mcmc,
  title={MCMC using Hamiltonian dynamics},
  author={Neal, Radford M and others},
  journal={Handbook of markov chain monte carlo},
  volume={2},
  number={11},
  pages={2},
  year={2011},
  publisher={Chapman and Hall/CRC}
}

@inproceedings{welling2011bayesian,
  title={Bayesian learning via stochastic gradient Langevin dynamics},
  author={Welling, Max and Teh, Yee W},
  booktitle={Proceedings of the 28th international conference on machine learning (ICML-11)},
  pages={681--688},
  year={2011}
}

@article{vollmer2015non,
  title={(Non-) asymptotic properties of stochastic gradient Langevin dynamics},
  author={Vollmer, Sebastian J and Zygalakis, Konstantinos C and others},
  journal={arXiv preprint arXiv:1501.00438},
  year={2015}
}

@article{song2019generative,
  title={Generative modeling by estimating gradients of the data distribution},
  author={Song, Yang and Ermon, Stefano},
  journal={Advances in neural information processing systems},
  volume={32},
  year={2019}
}

@article{ho2020denoising,
  title={Denoising diffusion probabilistic models},
  author={Ho, Jonathan and Jain, Ajay and Abbeel, Pieter},
  journal={Advances in neural information processing systems},
  volume={33},
  pages={6840--6851},
  year={2020}
}

@article{song2020score,
  title={Score-based generative modeling through stochastic differential equations},
  author={Song, Yang and Sohl-Dickstein, Jascha and Kingma, Diederik P and Kumar, Abhishek and Ermon, Stefano and Poole, Ben},
  journal={arXiv preprint arXiv:2011.13456},
  year={2020}
}

@article{dhariwal2021diffusion,
  title={Diffusion models beat gans on image synthesis},
  author={Dhariwal, Prafulla and Nichol, Alexander},
  journal={Advances in neural information processing systems},
  volume={34},
  pages={8780--8794},
  year={2021}
}

@inproceedings{aali2023solving,
  title={Solving inverse problems with score-based generative priors learned from noisy data},
  author={Aali, Asad and Arvinte, Marius and Kumar, Sidharth and Tamir, Jonathan I},
  booktitle={2023 57th Asilomar Conference on Signals, Systems, and Computers},
  pages={837--843},
  year={2023},
  organization={IEEE}
}

@article{ying2024feature,
  title={Feature selection as deep sequential generative learning},
  author={Ying, Wangyang and Wang, Dongjie and Chen, Haifeng and Fu, Yanjie},
  journal={ACM Transactions on Knowledge Discovery from Data},
  volume={18},
  number={9},
  pages={1--21},
  year={2024},
  publisher={ACM New York, NY}
}

@article{gong2025neuro,
  title={Neuro-symbolic embedding for short and effective feature selection via autoregressive generation},
  author={Gong, Nanxu and Ying, Wangyang and Wang, Dongjie and Fu, Yanjie},
  journal={ACM Transactions on Intelligent Systems and Technology},
  volume={16},
  number={2},
  pages={1--21},
  year={2025},
  publisher={ACM New York, NY}
}

@unknown{arun2025delta,
author = {Malarkkan, Arun and Bai, Haoyue and Kaushik, Anjali and Fu, Yanjie},
year = {2025},
month = {08},
pages = {},
title = {DELTA: Variational Disentangled Learning for Privacy-Preserving Data Reprogramming},
doi = {10.48550/arXiv.2509.00693}
}

@article{ying2025survey,
  title={A survey on data-centric ai: Tabular learning from reinforcement learning and generative ai perspective},
  author={Ying, Wangyang and Wei, Cong and Gong, Nanxu and Wang, Xinyuan and Bai, Haoyue and Malarkkan, Arun Vignesh and Dong, Sixun and Wang, Dongjie and Zhang, Denghui and Fu, Yanjie},
  journal={arXiv preprint arXiv:2502.08828},
  year={2025}
}

@inproceedings{malarkkan2024multi,
  title={Multi-view causal graph fusion based anomaly detection in cyber-physical infrastructures},
  author={Malarkkan, Arun Vignesh and Wang, Dongjie and Fu, Yanjie},
  booktitle={Proceedings of the 33rd ACM International Conference on Information and Knowledge Management},
  pages={4760--4767},
  year={2024}
}

@article{malarkkan2025incremental,
  title={Incremental Causal Graph Learning for Online Cyberattack Detection in Cyber-Physical Infrastructures},
  author={Malarkkan, Arun Vignesh and Wang, Dongjie and Bai, Haoyue and Fu, Yanjie},
  journal={IEEE Transactions on Big Data},
  year={2025},
  publisher={IEEE}
}

@article{malarkkan2025causal,
  title={Causal Graph Profiling via Structural Divergence for Robust Anomaly Detection in Cyber-Physical Systems},
  author={Malarkkan, Arun Vignesh and Bai, Haoyue and Wang, Dongjie and Fu, Yanjie},
  journal={arXiv preprint arXiv:2508.09504},
  year={2025}
}

@article{malarkkan2026causally,
  title={Causally-Guided Automated Feature Engineering with Multi-Agent Reinforcement Learning},
  author={Malarkkan, Arun Vignesh and Ying, Wangyang and Fu, Yanjie},
  journal={arXiv preprint arXiv:2602.16435},
  year={2026}
}

@article{malarkkan2025rethinking,
  title={Rethinking spatio-temporal anomaly detection: A vision for causality-driven cybersecurity},
  author={Malarkkan, Arun Vignesh and Bai, Haoyue and Wang, Xinyuan and Kaushik, Anjali and Wang, Dongjie and Fu, Yanjie},
  journal={arXiv preprint arXiv:2507.08177},
  year={2025}
}
